\documentclass[11pt]{article}
\usepackage{coling2020}
\usepackage{times}
\usepackage{latexsym}
\usepackage{graphicx}
\usepackage{amsmath}
\usepackage{amsfonts,amssymb}
\usepackage{subfigure}
\usepackage{url}
\usepackage{multirow}
\usepackage{enumitem}
\usepackage{pifont}
\usepackage{booktabs}

\usepackage[colorlinks,linkcolor=blue,anchorcolor=blue,citecolor=blue]{hyperref}
\newcommand{\cmarkb}{\ding{52}}
\newcommand{\xmark}{\ding{55}}

\usepackage{microtype}

\title{Molweni: A Challenge Multiparty Dialogue-based Machine Reading Comprehension Dataset with Discourse Structure}
 
\author{Jiaqi Li\textsuperscript{1}, Ming Liu\textsuperscript{1,3}, Min-Yen Kan\textsuperscript{2}, Zihao Zheng\textsuperscript{1}, Zekun Wang\textsuperscript{1} \\\textbf{Wenqiang Lei}\textsuperscript{\textbf{2}}, \textbf{Ting Liu}\textsuperscript{\textbf{1,3}}, \textbf{Bing Qin}\textsuperscript{\textbf{1,3}}\thanks{\ \ Corresponding author.}\\
1. Harbin Institute of Technology, Harbin, China. \\
2. National University of Singapore, Singapore. \\
3. Peng Cheng Laboratory, Shenzhen, China.\\
\{jqli, mliu, zhzheng,zkwang,tliu,qinb\}@ir.hit.edu.cn\\
kanmy@comp.nus.edu.sg, wenqianglei@gmail.com}

\date{}

\begin{document}

\maketitle

\begin{abstract}

Research into the area of multiparty dialog has grown considerably over recent years.  We present the \textit{Molweni} 
dataset\footnote{\url{https://github.com/HIT-SCIR/Molweni}}, a machine reading comprehension (MRC) dataset with discourse structure built over multiparty dialog. Molweni's source samples from the Ubuntu Chat Corpus, including 10,000 dialogs comprising 88,303 utterances.  We annotate 30,066 questions on this corpus, including both answerable and unanswerable questions.  Molweni also uniquely contributes discourse dependency annotations in a modified Segmented Discourse Representation Theory (SDRT; \cite{asher2016discourse}) style for all of its multiparty dialogs, contributing large-scale (78,245 annotated discourse relations) data to bear on the task of multiparty dialog discourse parsing. Our experiments show that Molweni is a challenging dataset for current MRC models: {\tt BERT-wwm}, a current, strong SQuAD~2.0 performer, achieves only 67.7\% $F_1$ on Molweni's questions, a 20+\% significant drop as compared against its SQuAD~2.0 performance.
\end{abstract}

\section{Introduction}

Research into multiparty dialog has recently grown considerably, partially due to the growing ubiquity of dialog agents. Multiparty dialog applications such as discourse parsing and meeting summarization are now mainstream research \cite{shi2019deep,GSNijcai2019,li2019keep,zhao2019abstractive,sun2019dream,N16IntegerLP,D15DP4MultiPparty}. Such applications must consider the more complex, graphical nature of discourse structure: coherence between adjacent utterances is not a given, unlike standard prose where sequential guarantees hold.  

In a separate vein, the area of machine reading comprehension (MRC) research has also made unbridled progress recently.  Most existing datasets for machine reading comprehension (MRC) adopt well-written prose passages and historical questions as inputs \cite{richardson2013mctest,rajpurkar2016squad,lai2017race,choi2018quac,reddy2019coqa}. 


Reading comprehension for dialog --- as the intersection of these two areas --- has naturally begun to attract interest. 
\newcite{ma2018challenging} constructed a small dataset for passage completion on multiparty dialog, but which has been easily dispatched by CNN+LSTM models using attention. The DREAM corpus \cite{sun2019dream} is an MRC dataset for dialog, but only features a minute fraction (1\%) of multiparty dialog. FriendsQA is a small-scale span-based MRC dataset for multiparty dialog, which derives from TV show \textit{Friends}, including 1,222 dialogs and 10,610 questions \cite{yang-choi-2019-friendsqa}. 
The limited number of dialogs in FriendsQA makes it infeasible to train more complex
model to represent multiparty dialogs due to overfitting, and the lack of annotated discourse structure prevents models from making full use of the characteristics of multiparty dialog.

Dialog-based MRC thus varies from other MRC variants in two key aspects:
\begin{enumerate}
	\item[C1.] Utterances of multiparty dialog are much less locally coherent than in prose passages.  A passage is a continuous text where there is a discourse relation between every two adjacent sentences. Therefore, we can regard each paragraph in a passage as a linear discourse structure text. In contrast, there may be no discourse relation between adjacent utterances in a multiparty dialog. As such, we regard the discourse structure of a multiparty dialog  as a dependency graph where each node is an utterance.
	
	\item[C2.] Multiparty dialog subsumes the special case of two-party dialog. 
	In most cases, the discourse structure of a two-party dialog is tree-like, where discourse relations mostly occur between adjacent utterances. However, in multiparty dialog, such assumptions hold less often as two utterances may participate in discourse relations, though they are very distant. 
\end{enumerate}



\begin{figure}
	\centering
	\includegraphics[width=\textwidth]{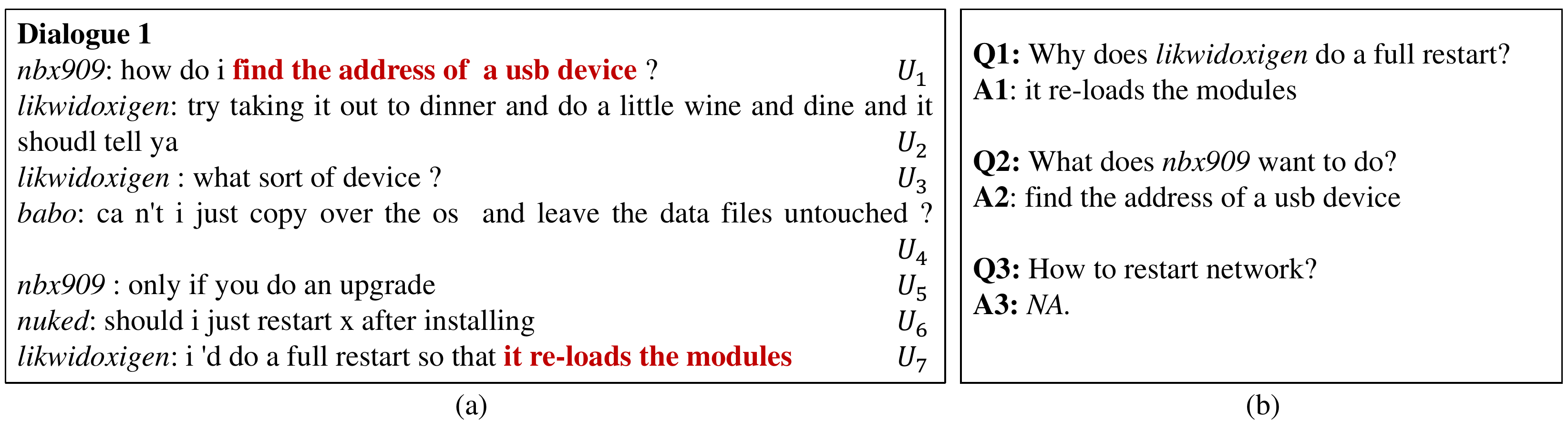}
	\caption{(Dialog~1) A corpus example from Molweni. There are four speakers in the dialog: \textit{nbx909}, \textit{Likwidoxigen}, \textit{babo}, and \textit{nuked}. In total, the speakers make seven utterances: $U_1$ to $U_7$. Our annotators proposed three questions against the provided dialog: Q1--3, where Q1 and Q2 are answerable questions, and Q3 is unanswerable. Due to the properties of informal dialog, the instances in our corpus often have grammatical errors.}
	\label{fig:framework}
\end{figure}

Prior MRC works do not consider the properties of multiparty dialog. To address this gap in understanding of multiparty dialog, we created Molweni. In Dialog~1 (\textit{cf.} Fig~1), four speakers converse over seven utterances.  We additionally employ annotators to read the passage and contribute questions: in the example, the annotators propose three questions: two answerable and one unanswerable.  We observe that adjacent utterance pairs can be incoherent, illustrating the key challenge.  It is non-trivial to detect discourse relations between non-adjacent utterances; and crucially, difficult to correctly interpret a multiparty dialog without a proper understanding of the input's complex structure.

We derived Molweni from the large-scale multiparty dialog Ubuntu Chat Corpus \cite{lowe2015ubuntu}.  We chose the name {\it Molweni}, as it is the plural form of ``Hello'' in the Xhosa language, representing multiparty dialog in the same language as {\it Ubuntu}. Our dataset contains 10,000 dialogs with 88,303 utterances and 30,066 questions including answerable and unanswerable questions. All answerable questions are extractive questions whose answer is a span in the source dialog. For unanswerable questions, we annotate their plausible answers from dialog. Most questions in Molweni are 5W1H questions -- {\it Why}, {\it What}, {\it Who}, {\it Where}, {\it When}, and {\it How}. For each dialog in the corpus, annotators propose three questions and find the answer span (if answerable) in the input dialog. 

To assess the difficulty of Molweni as an MRC corpus, we train BERT's whole word masking model on Molweni, achieving a 54.7\% exact match (EM) and 67.7\% $F_1$ scores. Both scores show larger than 10\% gap with 
human performance, validating its difficulty. Due to the complex structure of multiparty dialog,  human performance just achieves 80.2\% $F_1$ on Molweni. 
In particular, annotators agreed that knowledge of the correct discourse structure would be helpful for systems to achieve better MRC performance.

This comes to the second key contribution of Molweni.  We further annotated all 78,245 discourse relations in all of Molweni's dialogs, in light of the potential help that annotated discourse structure might serve.  Prior to Molweni, the STAC corpus is the only dataset for multiparty dialog discourse parsing \cite{asher2016discourse}.  However, its limited scale (only 1K dialogs) disallow data-driven approaches to discourse parsing for multiparty dialog. We saw the additional opportunity to empower and drive this direction of research for multiparty dialog processing.





\section{Related work}


\paragraph{Discourse parsing for multiparty dialog.} Prior to Molweni, STAC was the only corpus containing annotations for discourse parsing on multiparty chat dialogs \cite{asher2016discourse}. The corpus derives from the online version of the game \textit{The Settlers of Catan}. The game is a multiparty, win--lose game. We introduce the senses of discourse relation in STAC in Section~3.2. The STAC corpus contains 1,091 dialogs with 10,677 utterances and 11,348 discourse relations. Compared with STAC, our Molweni dataset contains 10,000 dialogs comprising 88,303 utterances and 78,245 discourse relations.

\paragraph{Machine reading comprehension.} 
There are several types of datasets for machine comprehension, including multiple-choice datasets \cite{richardson2013mctest,lai2017race}, answer sentence selection datasets \cite{wang2007jeopardy,yang2015wikiqa} and extractive datasets \cite{rajpurkar2016squad,joshi2017triviaqa,trischler2017newsqa,rajpurkar2018know} . 
To extend existing corpora, our Molweni dataset is constructed to be an extractive MRC dataset for multiparty dialog, which includes both answerable questions and unanswerable questions. Similar to Squad~2.0 \cite{rajpurkar2018know}, we also annotate plausible answers for unanswerable questions. Three closely related datasets are 
the extended crowdsourced {\it Friends} corpus \cite{ma2018challenging}, DREAM \cite{sun2019dream} and FriendsQA \cite{yang-choi-2019-friendsqa}. Different from these three MRC datasets for dialog, Molweni contributes the discourse structure of dialogs, and additional instances of multiparty dialogs and unanswerable questions.




\section{The Molweni corpus}



Our dataset derives from the large scale multiparty dialogs dataset --- the Ubuntu Chat Corpus \cite{lowe2015ubuntu}. 
We list our three reasons in choosing the Ubuntu Chat Corpus as the base corpus for annotation.
\begin{itemize}
	\item First, the Ubuntu dataset is a large multiparty dataset. 
	After filtering the dataset by only retaining all utterances with response relations, there are still over 380K sessions and 1.75M utterances. In each session, there are 3-10 utterances and 2-7 interlocutors.
	\item Second, it is easy to annotate the Ubuntu dataset. The Ubuntu dataset already contains Response-to relations that are discourse relations between different speakers' utterances. For annotating discourse dependencies in dialog, we only need to annotate relations between the same speaker's utterances and the specific sense of discourse relation. Because the length of dialogs in the Ubuntu dataset is not too long, we can easily summarize dialogs and propose some questions for the dialog.
	\item Third, there are many papers doing experiments on the Ubuntu dataset, and the dataset has been widely recognized. For example, \newcite{kummerfeld2019large} proposed a large-scale, offshoot dataset for conversation disentanglement based on the Ubuntu IRC log.
	Also recently, \newcite{GSNijcai2019} also used the Ubuntu Chat Corpus as their dataset for learning dialog graph representation. 
\end{itemize}

The discourse dependency structure of each multiparty dialog can be regarded as a discourse dependency graph where each node is an utterance. To learn better graph representation of multiparty dialogs, we filter the Ubuntu Chat Corpus for complex dialogs -- those dialogs with 8--15 utterances and 2--9 speakers.  
As multiparty dialog is already intensely complex for the current state of the art, we chose to further simplify in our selection criteria, additionally filtering out dialogs with long utterances (more than 20 words).
Finally, we 
randomly chose a subset of $\sim$10,000 dialogs with 88,303 utterances from the Ubuntu dataset. 
We give an overview of Molweni's key demographics in Table~1.


\begin{table}[th]
    \begin{center}
    \caption{\label{font-table} Overview of Molweni for MRC.}
        \begin{tabular}{l|l|c|c|c}
            \hline \bf   \bf   & \bf Train  & \bf Dev & \bf Test & \bf Total \\ 
            \hline
            Number of Dialogs      & 8,771	   &  883 & 100 & 9,754 \\
            Number of Utterances     & 77,374   &  7,823 &  845 & 86,042\\ 
            Number of Questions     & 24,682 & 2,513 & 2,871 & 30,066  \\
            \hline
        \end{tabular}
    \end{center}
    \vspace{-0.16in}
\end{table}

10,000 dialogs are divided into two parts: 100 dialogs in common (public dialog) and 9,900 dialogs for different annotators (private dialog). Each annotator is asked to annotate 1,090 dialogs (990 private dialogs and 100 public dialogs)  in two aspects: machine reading comprehension and discourse structure. All annotators chose to annotate the discourse structure of the dialog, and then propose questions and find answer spans for the dialog. 
All annotators agreed that it would be helpful to annotate the MRC task after annotating the discourse structure. 

In total, our subjects annotated 9,754 dialogs, 
slightly fewer than 10,000 dialogs,
consisting of 88,303 utterances, and contributed 30,066 questions for machine reading comprehension and 78,245 discourse relation annotations.
There are 8,771 dialogs in the demarcated {\sc Train} set for both machine reading comprehension and discourse parsing tasks. 883 dialogs are used for {\sc Dev} set. Each annotator is asked to propose three questions per dialog. There are 100 dialogs in common for all ten annotators, and these 100 dialogs comprise our {\sc Test} set. Each dialog in the training set and develop set has three questions. 
Our annotation team proposed a total of 2,871 questions for the 100 dialogs in {\sc Test} sets. 

Detailed statistics are shown in Table~2. The average number of speakers per dialog is 3.51, which means that most dialogs are multiparty (as opposed to 2-party) dialogs. 
The number of two-party dialogs and multiparty dialogs in our dataset is 2,117 and 7,883, respectively. 
In Molweni, the average and maximum length of the selected dialogs are 8.82 and 14 utterances, respectively, and the number of answerable and unanswerable questions are 25,779 and 4,287, respectively.

\begin{table}[th]
\vspace{-0.10in}
    \begin{center}
        \caption{\label{font-table} Detailed statistics for the {\it Molweni} corpus.}
        \begin{tabular}{p{10cm}|rl}
            \hline \bf   \bf Metric  & \multicolumn{2}{c}{\bf Number}   \\ 
            \hline
            Average / Maximum number of speakers per dialog   & 3.51 & / 9      \\ 
            Average / Maximum question length (in tokens) & 5.91 & / 18 \\
            Average / Maximum answer length (in tokens) & 4.08 & / 19 \\
            Average / Maximum dialogue length (in tokens) & 104.4 & / 208 \\
            Average / Maximum dialogue length (in utterances) & 8.82 & / 14 \\
            Vocabulary size & \multicolumn{2}{c}{24,615} \\
            Answerable questions &  \multicolumn{2}{c}{25,779} \\
            Unanswerable questions &  \multicolumn{2}{c}{4,287}\\
            \hline
        \end{tabular}
    \end{center}
    
\end{table}

\begin{table*}[t]
	\vspace{-0.30in}
	\caption{\label{font-table} Comparison of Molweni with other MRC datasets on answer type, text type (dialogue or written text), multiparty dialogs or not, unanswerable questions, and discourse structure.}

	\centering
		\setlength{\tabcolsep}{1mm}{
			\resizebox{\textwidth}{!}{
			\begin{tabular}{l | c | c | c | c | c}
				\hline 
				\multirow{2}{*}{\bf Dataset} & \bf Answer  &\bf Dialogue   & \bf Multiparty  &  \bf Unanswerable  & \bf Discourse \\ 
				& \bf type & \textbf{text} & \textbf{dialogues} & \textbf{questions} & \textbf{structure} \\
				\hline
				
				\hline
				
				RACE \cite{lai2017race} & multiple-choice & \xmark  & \xmark   & \xmark & \xmark \\
				NarrativeQA \cite{kovcisky2018narrativeqa} & abstractive & \xmark & \xmark  & \xmark & \xmark\\
				CoQA \cite{choi2018quac} & abstractive & \xmark & \xmark  & \cmarkb & \xmark \\
				SQuAD 2.0 \cite{rajpurkar2018know} & extractive & \xmark  &  \xmark  & \cmarkb & \xmark\\
				QuAC \cite{choi2018quac} & extractive & \xmark  & \xmark   & \cmarkb & \xmark \\
				\cite{ma2018challenging} & cloze & \cmarkb  & \cmarkb  & \xmark & \xmark \\
				DREAM \cite{sun2019dream} & multiple-choice & \cmarkb & \cmarkb   &  \xmark & \xmark \\
				FriendsQA \cite{yang-choi-2019-friendsqa} & extractive & \cmarkb & \cmarkb   &  \xmark & \xmark \\
				\hline
				
				Molweni (Our) & extractive &  \cmarkb &  \cmarkb   &  \cmarkb &  \cmarkb\\
				\hline
				
				\hline
		\end{tabular}}}

\end{table*}


\subsection{Annotation for machine reading comprehension}

We hired ten annotators to construct our Molweni dataset. As the Ubuntu corpus is technical in nature, all annotators are undergraduate students whose major is computer science to annotate the corpus. Annotators are non-native English speakers but who have an English proficiency certificate. They are all familiar with Linux operation system.

Annotators propose three questions for each dialog and annotate the span of answers in the input dialog. 
There are two types of questions in our corpus, namely, answerable questions and unanswerable questions:
\begin{enumerate}
	\item \textbf{Answerable questions.} For these questions, the answer is a continuous span from source dialog. Annotators were asked to label answers from input dialog and ensure answers were succinct, without including extraneous text. 
	\item \textbf{Unanswerable questions.} To make the reading comprehension task more challenging, we annotate unanswerable questions and their plausible answers (PA). The plausible answers are quite related to unanswerable questions.
\end{enumerate}

We compare Molweni against other datasets in Table~3. We see that existing dialog MRC datasets neither contribute either unanswerable questions, nor annotated discourse structure. Due to the complex structure of multiparty dialogs, we believe that it is essential to adopt the discourse dependency structure for the machine modeling towards multiparty dialog understanding. To the best of our knowledge to date, Molweni is the only MRC dataset that is annotated with discourse structure.

We give example questions from Molweni in Table~4. In particular, most of the questions in our dataset are questions lead by {\it Why}, {\it What}, {\it Who}, {\it Where}, {\it When}, and {\it How}. Only a small proportion of the questions are Other questions; questions lead by words such as {\it Do}, {\it Which}, and {\it Whose}. When annotators propose questions, they are asked to consider the characteristics of multiparty dialogs. For example, for {\it Why} and {\it How} questions, it is essential to know the question--answer pair and the cause--result in the dialog. For {\it How} questions, it is important to understand the role of speakers in order to properly represent the multiparty dialog. As such, {\it Why} and {\it How} often require a deeper understanding of the dialog.

        
        

\begin{table}[thbp]
\vspace{-0.20in}
  \centering
  \caption{Examples of questions in Molweni.}
  \resizebox{\textwidth}{!}{
    \begin{tabular}{l|ll|c}
    \toprule
    Question  & \multicolumn{2}{c}{Example} & \multicolumn{1}{|l}{Proportion(\%)} \\
    \midrule
    {\it How} & How to do an upgrade? & How can I use this machine? & 9.9 \\
    {\it Why} & Why is it not mounted? & Why does \textit{jimcoonact} meet the error? & 4.3 \\
    {\it Who}  & Who is chart's service customers? & Who is using ubuntu? & 4.7 \\
    {\it When} & When does \textit{rhodry} have the error? & When is \textit{SuperMiguel} back? & 1.7 \\
    {\it Where} & Where did \textit{earthen} write in? & Where is the device? & 5.7 \\
    {\it What}  & What does \textit{elnomade} choose? & What does \textit{noone} need? & 71.7 \\
    Others & Does \textit{elnomade} choose the print? & Which version does \textit{xxiao} find? & 1.9
    \\
    \bottomrule
    \end{tabular}}%
  \label{tab:addlabel}%
\end{table}%

\subsection{Annotation for discourse structure of multiparty dialogs}

The task of discourse parsing for multiparty dialogs is to determine the discourse relations among utterances.  To enable better future modeling of  such multiparty discourse, we represent a multiparty dialog by a directed acyclic graph (DAG). The process of annotating the discourse structure consists of two parts: predicting the links between utterances, and classifying the sense of the resultant discourse relation. Table~5 gives an overview of the statistics for Molweni's discourse parsing annotations.

\begin{table}[t]
    \vspace{-0.20in}
    \begin{center}
    \caption{\label{font-table} Statistics of Molweni's annotated discourse relations.}
        \begin{tabular}{l|l|c|c|c}
            \hline \bf   \bf   & \bf Train  & \bf Dev & \bf Test & \bf Total \\ 
            \hline
            Number of Dialogs      & 9,000	   &  500 & 500 & 10,000 \\
            Number of Utterances     & 79,487   &  4,386 &  4,430 & 88,303\\ 
            Number of Relations     & 70,454    & 3,880 & 3,911 & 78,245  \\
            \hline
        \end{tabular}
    \end{center}
    \vspace{-0.16in}
\end{table}

An edge between two utterances represents the existence of a discourse dependency relation. The direction of the edge represents the direction of discourse dependency. In this subtask, what annotators need to do is to confirm whether two utterances have a discourse relation. Following the convention in the Penn Discourse Treebank (PDTB) \cite{prasad2008penn}, we term the two utterances as \textit{Arg1} and \textit{Arg2}, 
sequentially. 

\begin{table}[htbp]
  \centering
  \caption{Discourse relation types in Molweni and their meanings, listed in order of descending frequency in the corpus.}
  \resizebox{\textwidth}{!}{
    \begin{tabular}{l|l|c}
    \toprule
    Relation & \multicolumn{1}{c}{Meaning} & \multicolumn{1}{|l}{Proportion(\%)} \\
    \midrule
    Comment  & Arg2 comments Arg1.   & 31.7 \\
    Clarification\_question  & Arg2 clarifies Arg1. & 24.0 \\
    Question-answer\_pair  & Arg1 is a question and Arg2 is the answer of Arg1. & 20.1 \\
    Continuation  & Arg2 is the continuation of Arg1. & 6.7 \\
    Acknowledgement  & Arg2 acknowledges Arg1. & 3.2 \\
    Q-Elab  & Arg1 is a question and Arg2 tries to elaborate Arg1. & 3.0 \\
    Result  & Arg2 is the effect brought about by the situation described in Arg1.   & 2.6 \\
    Elaboration  & Arg2 elaborates Arg1. & 2.2 \\
    Explanation  & Arg2 is the explanation of Arg1. & 1.6 \\
    Correction  & Arg2 corrects Arg1.   & 1.2 \\
    Contrast  & Arg1 and Arg2 share a predicate or property and a difference on shared property.  & 1.2 \\
    Conditional  & Arg1 is the condition of Arg2 or Arg2 is the condition of Arg1. & 1.0 \\
    Background  & Arg2 is the background of Arg1. & 0.4 \\
    Narration  & Arg2 is the narration of Arg1. & 0.3 \\
    Alternation  & Arg1 and Arg2 denote alternative situations. & 0.2 \\
    Parallel  & Arg2 and Arg1 are parallel and present almost the same meaning. & 0.2 \\
    \bottomrule
    \end{tabular}}%
  \label{tab:addlabel}%
\end{table}%


Discourse relations on two non-adjacent utterances are rare in prose but common in multiparty dialogs. When we annotate dialogs, annotators should sequentially read the dialog from its beginning  to its final utterance. For each utterance, annotators need to find at least one parent node from among the previous utterances. We assume that the discourse structure is a connected graph and no utterance is isolated. 

After we find the discourse relation between two utterances, we then need to confirm the specific relation sense. Although the sense hierarchy in PDTB has been broadly adopted~\cite{lei2017swim,lei2018linguistic}, we adopt the modified Segmented Discourse Reprsentation Theory (SDRT) hierarchy, defined in STAC dataset \cite{asher2016discourse}, as it is designed specifically for multiparty dialog. There are 16 discourse relations in the STAC schema, as given in Table~6, where the top four most frequent relations (Comment, Clarification Question, Question Answer Pair (QAP), and Continuation) make up over 80\% of the relations in the corpus.

For the discourse parsing task, we used 500 dialogs for development and 500 dialogs for testing which is different from the MRC tasks.  In opposition to the frequent relationships, there are also four types of relations that individually account for less than one percent of the corpus, namely, Alternation, Background, Narration, and Parallel. This is similar to the proportion of these four types of relations in the STAC dataset as well: only 0.5--2.0\%. Next, according to the distribution of all kinds of relations, we need to consider merging some rare relation types in future work, so as to propose a more practical sense hierarchy for multiparty dialogs. 



Multi-relational link prediction aims to predict missing links in an edge-labeled graph.  This task focuses on the relations between entities \cite{bordes2013translating}. However, discourse parsing focus on finding the discourse dependency arcs between different utterances. 


The discourse dependency structures of Dialog~1 and Dialog~2 are shown in Fig.~3 where each utterance is represented as a node in the dependency graph. The label on the link in the discourse dependency relation.  Dialog~1 ({\it cf.} Table~1) is a multiparty dialog with four speakers: \textit{nbx909},\textit{likwidoxigen}, \textit{babo}, \textit{nuked} and seven utterances. Dialog~2 (\textit{cf.} Fig.~2) has two speakers: \textit{toma-} and \textit{woodgrain}, and eight utterances in total. From Fig.~1, we can find that most of the discourse relations of two-party dialog occurs between adjacent utterances.


\begin{figure}
	\centering
	\includegraphics[width=\textwidth]{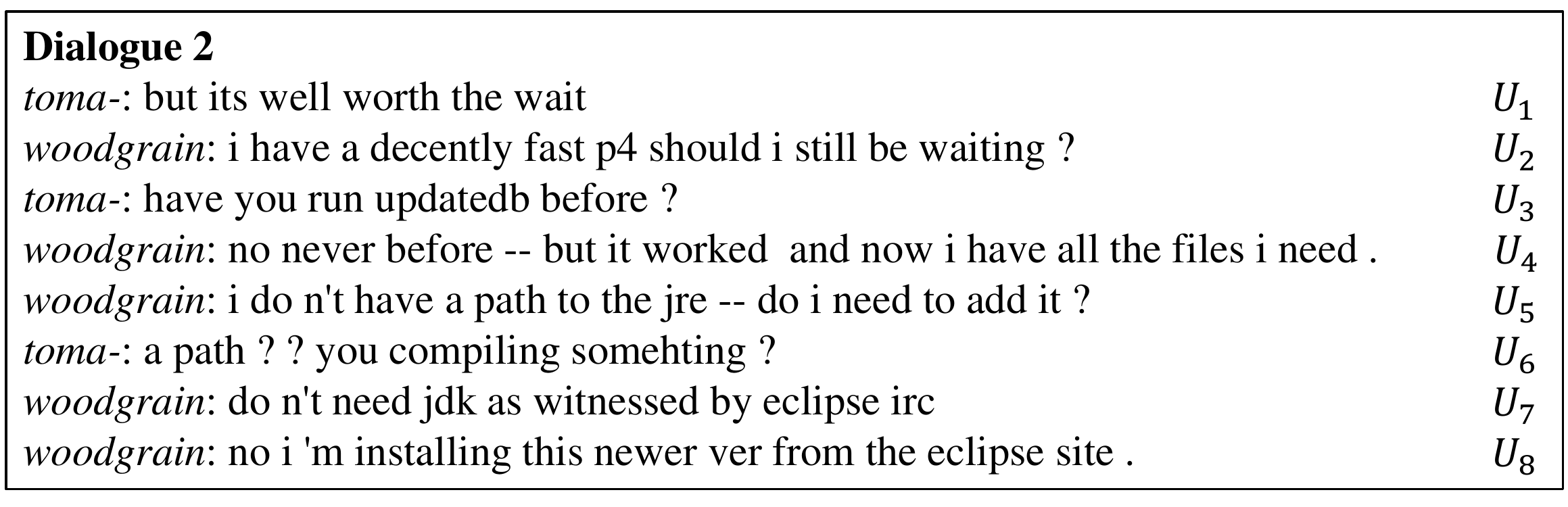}
	\caption{Dialog~2 is a two-party dialog example with eight utterances --- $U_1$ to $U_8$ --- proposed by two speakers: \textit{toma-} and \textit{woodgrain}.}
	\label{fig:framework}
\end{figure}

\begin{figure}
	\centering
	\includegraphics[width=\textwidth]{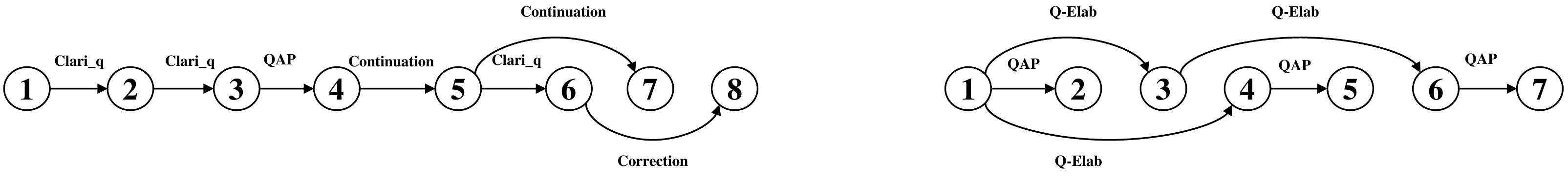}
	\caption{The discourse dependency structure and relations for Dialog~2 (Left, two-party) and Dialogue~1 (Right, multiparty). Clari\_q, QAP, and Q-Elab are respectively short for Clarification\_question, Question-answer\_pair, and Question-Elaboration. The label on the link represents the discourse dependency relations between two utterances.}
	\label{fig:framework}
\end{figure}



\subsection{Data Quality}

To ensure the quality of the corpus, we adopt two ways to check the annotation: a manual, human check as well as a programmatic check. 
\begin{itemize}
	\item {\bf Manually Check}. Two authors of our Molweni dataset sample some instances to check the quality of proposed questions and feedback bad questions to the annotator.
	\item {\bf Programmatic Check}. If answers cannot be found in the source dialog, the annotator would be asked to annotate the dialogs again until passing the check. We additionally check that the questions are grammatically correct using the \textit{Grammarly} web application \footnote{\url{https://app.grammarly.com/}}.
\end{itemize}
After four rounds of revision, we obtain the currently published version of the dataset.

We calculate the Fleiss Kappa value to check on the interannotator consistency.  A Kappa value of 1.0 signifies complete agreement, and a value of 0.0 signifies completely uncorrelated judgments.  Kappa values above  The Kappa value of discourse dependency links is 0.91 which is an almost perfect agreement because the Ubuntu dataset initially contains the response-to relations, and annotators adopt most of the links. The final Kappa value of both links and relations is 0.56 among annotators, close to that of 0.58 obtained in the original STAC corpus.  
One reason for the drop of Kappa after labeling relation types is the discourse relation recognition is a multi-label task. There could be more one relation between two utterances in a dialog, which would easily make ambiguities.

\section{Experiments}

We now introduce our baseline experiments on our dataset. We consider the two tasks of discourse parsing and machine comprehension for multiparty dialog.

\subsection{Machine reading comprehension for multiparty dialogs}

\paragraph{Methods.} SQuAD~2.0 is an MRC dataset that adopts a passage as the input and the answer is a span from input passage \cite{rajpurkar2018know}.  We adopt the following existing methods for SQuAD~2.0 on our dataset. In this paper, we use three different kinds of settings of BERT: BERT-base, BERT-large, and BERT-whole word masking (BERT-wwm). 
We concatenate all utterances from input dialog as a passage, and each utterance includes speaker and text. 
We used the open-source code of BERT to perform our experiments\footnote{\url{https://github.com/google-research/bert}}.


BERT is a bidirectional encoder from  transformers \cite{devlin2019bert}. To learn better representations for text, BERT adopts two objectives: masked language modeling and the next sentence prediction during pretraining.  In the BERT-wwm, if a part of a complete word WordPiece is replaced by [mask], the other parts of the same word will also be replaced by mask, which is the whole word mask. 

\begin{itemize}
{
    \item \textbf{BERT-base:} 12-layer, 768-hidden, 12-heads, 110M parameters.
    \item \textbf{BERT-large:} 24-layer, 1024-hidden, 16-heads, 340M parameters. The difference between BERT-base and BERT-large is in the number of the parameters; there is no difference in model architecture.
    \item \textbf{BERT-wwm:} 24-layer, 1024-hidden, 16-heads, 340M parameters. The original word segmentation method based on WordPiece segments a complete word into several affixes. When generating training samples, these separated affixes are randomly replaced by {\tt [mask]}. 
   
        
}
\end{itemize}

\begin{table*}
\caption{\label{font-table} Results of machine reading comprehension for multiparty dialogues.}
\begin{center}
\begin{tabular}{c||c|c||c|c}
\hline
\multirow{2}{*}{\textbf{Method}}&
\multicolumn{2}{c||}{\textbf{EM}}& 
\multicolumn{2}{c}{\textbf{F1}} \\
\cline{2-5}
  & \textbf{Squad 2.0} & \textbf{Our} & \textbf{Squad 2.0} & \textbf{Our} \\
\hline
BERT-base & 73.1 & 45.3 & 76.2 & 58.0\\
BERT-large & 80.0 & 51.8 & 83.1 & 65.5 \\
BERT-wwm & 86.7 & 54.7 & 89.1 & 67.7 \\
\hline
Human performance & 86.8 & 64.3 & 89.4 & 80.2 \\
Human-machine gap & 0.1 & 9.6 & 0.3 & 12.5 \\
\hline
\end{tabular}
\end{center}

\end{table*}


\paragraph{Evaluation Metric.} As our task is quite related to SQuAD~2.0, we adopt the same evaluation metrics: exact match (EM) and $F_1$ score to evaluate experiments. EM measures the percentage of predictions that match all words of the ground truth answers exactly. $F_1$ scores are a looser interpretation of match, measuring the average overlap between predictions and the ground truth answer. The results of machine reading comprehension for multiparty dialogs is shown in Table~7. 

\paragraph{Human upper bound.} We enlist two non-annotator volunteers whose majors are computer science to answer questions in the {\sc Test} set.  From Table~7, they achieved 64.3\% in EM and 80.2\% in $F_1$. This result show that (1) People can get good results in $F_1$, and (2) it is challenging to detect the accurate boundary of answers.
The results of humans show the challenge of machine comprehension for multiparty dialogs because the structure of a multiparty dialog is very complex and the language style in dialogs is very informal compared with well-written passage text.

\paragraph{Results.} For three BERT models, the BERT-wwm model achieves the best results on both SQuAD~2.0 and our Molweni dataset, followed by BERT-large and BERT-base. Especially, the BERT-wwm model gets 89.1\% $F_1$ score on SQuAD~2.0, very close to human performance.  The performance gap between BERT-wwm and human are 0.1\% EM and 0.3\% $F_1$ on SQuAD~2.0. However, on Molweni, BERT-wwm achieves only 67.7\% $F_1$, which has a significant large 12.5\% performance gap with human performance.

\paragraph{Case study}In this part, we will analyze the reason why BERT-wwm does not perform as well as it does on SQuAD~2.0. Fig.4 shows an example of dialog~3 in our Molweni test set with two bad cases of the BERT-wwm model. In Dialog~3, there are three speakers and ten utterances. The first question Q1 is about the user that asked for the address. The answer to BERT-wwm of Q1 is \textit{likwidoxigen}, but the gold answer is \textit{nbx909}. The second question Q2 is about the status of printers, but the model answers the status of people who makes the printers.

\begin{figure}
    \centering
    \includegraphics[width=\textwidth]{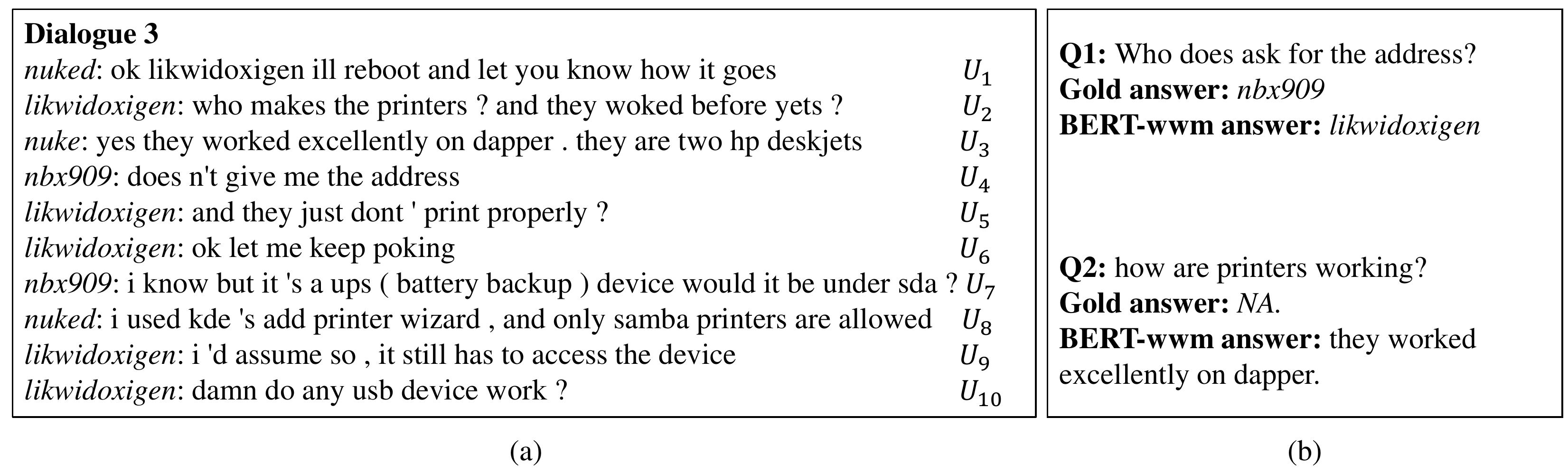}
    \caption{Dialogue3. (a) A real example from Molweni dataset with three speakers and ten utterances. (b) Two questions for Dialog~3 and the pridected answers of BERT-wwm model.}
    \label{fig:framework}
\end{figure}

We concatenate all utterances as the input which doesn't highlight the speaker information of the utterance. For Q1, after concatenating all utterances, \textit{likwidoxigen} would be the closest speaker in the input with the word 'address'. The speaker of utterances is the essential information for better understanding dialogs. 

On the other hand, when concatenating all utterances, the language model could automatically model the coherence between two adjacent utterances. But there could be no coherence between adjacent utterances, and the discourse structure of a multiparty dialog should not be regarded as a sequence but a graph. In most cases, every node (utterance) in the discourse dependency graph only has one parent node.

\subsection{Discourse parsing for multiparty dialogs}

\paragraph{Methods} We perform the Deep Sequential model on our Molweni corpus which is the state-of-the-art model on STAC. \newcite{shi2019deep} proposed the deep sequential model for discourse parsing on multiparty chat dialogs which adopted an iterative algorithm to learn the structured representation and highlight the speaker information in the dialog. The model jointly and alternately learns the dependency structure and discourse relations.

In this paper, we adopt two different kinds of the setting of the Deep Sequential model.
\begin{itemize}
{
    \item \textbf{Deep sequential} This is the original deep sequential model.  
    \item \textbf{Deep sequential(C)}  Considering that we adopt the same discourse relation hierarchy with the STAC corpus, we combine the training sets of STAC and Molweni as the training set for this model, we respectively test the model on STAC and Molweni.
}
\end{itemize}

\paragraph{Results} We adopt the F1 score to evaluate both links prediction and relation classification tasks, which is the same as previous literature. The results of discourse parsing for multiparty dialogs are shown in Table~8. For link prediction, we achieved higher results than the deep sequential model performed on STAC. On the other hand, we achieve comparable results for relations classification compared with STAC. After combining the training set of Molweni, the deep sequential model achieves better results on STAC which means the Molweni dataset can be beneficial to predict discourse dependency links.



\begin{table}
\caption{\label{font-table} Results of discourse parsing on multiparty dialogs ($F_1$-score). Deep sequential (C) means combine the training set of STAC and Molweni as the training set and test the model respectively.}
\begin{center}
\setlength{\tabcolsep}{0.1mm}{
\begin{tabular}{c||c|c||c|c}
\hline
\multirow{2}{*}{\textbf{Method}}&
\multicolumn{2}{c||}{\textbf{Link}}& 
\multicolumn{2}{c}{\textbf{Link \& Relation}} \\
\cline{2-5}
  & \textbf{STAC} & \textbf{Our} & \textbf{STAC} & \textbf{Our} \\
\hline
Deep sequential & 73.2 & 78.1 & 55.7 & 54.8\\
Deep sequential(C)  & 78.0 & 77.0 & 54.7 & 54.3\\
\hline
\end{tabular}}
\end{center}

\vspace{-0.10in}
\end{table}

\section{Conclusion}

In this paper, we introduce Molweni, a multiparty dialog dataset for machine reading comprehension (MRC). Compared with traditional textual structure, the dialog is concatenated by the utterances from multiple participants. We believe that discourse structure can provide potential help for understanding the dialog.  Therefore, we ask annotators to label the discourse dependency structure of the multiparty dialog and propose questions for the dialog. Annotation on a large number of dialog shows that tagging discourse structure can significantly help taggers understand dialog and raise higher quality questions. In the future, we will try to propose novel discourse parsing models for multiparty dialog and apply discourse structure in the reading comprehension task of multiparty dialog. 

\section*{Acknowledgements}

We thank anonymous reviewers for their helpful comments. Thanks to Yibo Sun, Tiantian Jiang, Daxing Zhang, Rongtian Bian, Heng Zhang, Hu Zhenyu and other students of Harbin Institute of Technology for their support in the process of data set annotation. Thanks to Wenpeng Hu, a Ph.D. student of Peking University, for providing preprocessed Ubuntu data. The research in this article is supported by the Science and Technology Innovation 2030 - "New Generation Artificial Intelligence" Major Project (2018AA0101901), the National Key Research and Development Project (2018YFB1005103), the National Science Foundation of China (61772156, 61976073) and the Foundation of Heilongjiang Province (F2018013).

\bibliographystyle{acl}
\bibliography{molweni}

\end{document}